\pdfoutput=1

\documentclass[11pt]{article}
\usepackage{acl}

\usepackage{times}
\usepackage{latexsym}

\usepackage[T1]{fontenc}

\usepackage[utf8]{inputenc}

\usepackage{microtype}

\usepackage{amsmath}
\usepackage{amssymb}
\usepackage{textcomp}
\usepackage{algorithm}
\usepackage{algpseudocode}
\usepackage{booktabs}
\usepackage{threeparttable}
\usepackage{multirow}
\usepackage{graphicx}
\usepackage{tabularx}

\title{
    {\Huge Technical Report} \\\vskip 3ex
    Competition Solution for Prompt Tuning using Pretrained Language Model
}

\author{
    Jiang-Long Song \and
    Wu-He Zou \and
    Feng Li \and
    Xiao-Lei Qin \and
    Wei-Dong Zhang \\
    Netease Games AI Lab, HangZhou, China \\
    \texttt{delta.z@aliyun.com}~~~
    \texttt{\{zouwuhe,lifeng06,qinxiaolei\}@corp.netease.com} \\
    \texttt{wdong.zhang@gmail.com}
}

\begin{document}
\maketitle

\begin{abstract}

Prompt tuning recently becomes a hot-spot in the applications of large pretrained language models on specific downstream tasks.
Regarding the Language Model as a Service (LMaaS), black-box tuning using derivative-free optimization (DFO) provides a novel approach to expand the practical scenarios of pretrained models and enrich the researches of few-shot learning.
In this report, we present our solution in this competition that is based on the LMaaS scenario.
Our solution consists of several modifications to BBTv2 \citep{sun2022bbtv2}, including multiple label words, selection of P0, rolling update strategy, multi-task loss from MLP classifier, and finally using the ensemble method to further improve generalization ability.
We also shared some strategies that we tried but didn't use in the final submission for further discussion.
In the end we raised a question about the SNLI dataset and the impact on the results, as well as our concerns about the competition.

\end{abstract}

\newpage
\tableofcontents

\onecolumn
\section{Background}

Pre-trained language models have greatly promoted the development of natural language processing in recent years.
As the scale of pre-training language models increases, it is found that good performance can be achieved in many tasks with few-shot learning.
However, due to the commercial considerations and potential risk of misuse, some large-scale language models are not open-sourced, but are released as a service through black-box APIs.
This scenario is called ``Language-Model-as-a-Service'' (LMaaS) \cite{Sun2022BlackBoxTF}.
In this scenario, users cannot get the parameters and gradients of the pre-trained language model, but can only get the inference results of APIs.
Therefore, how to improve NLP tasks only by inference results of APIs has become an important research direction.
This competition held by PaZhou Lab is designed to tune continuous and discrete prompts, requiring participants to optimize several few-shot learning tasks while only using the inference results of the pre-trained language model \footnote{\url{https://www.cvmart.net/race/10344/base}}.
The primary request is formulated as below.
\begin{equation}
    p^\ast = \arg \min_p \mathcal{L}(f(p, \mathbf{x}), y)
\end{equation}
where $p$ is the prompts to be optimized, $\mathcal{L}(\cdot)$ is a kind of loss function.
Basic statistics of five tasks in final stage of this competition and input templates we used are shown in Table \ref{tab:task}.

\begin{table}[!htb]
\centering
\caption{Basic statistics of five few-shot learning tasks}
\begin{tabular}{c c c l c}
    \toprule
    \multirow{2}{*}{Task} & \multirow{2}{*}{Classes} & Input & \multirow{2}{*}{Template} & Training/Dev/Test  \\
    & & Type & & Examples Per Seed \\
    \midrule
    DBPedia & 14 & single & \texttt{<prompt>~.~\{text\}~.~It was <mask>~.}   & 112/112/1792 \\
    QNLI    & 2  & pair   & \texttt{<prompt>~.~\{text1\}~<mask>~,~\{text2\}} &  32/32/1024  \\
    QQP     & 2  & pair   & \texttt{<prompt>~.~\{text1\}~<mask>~,~\{text2\}} &  32/32/1024  \\
    SNLI    & 3  & pair   & \texttt{<prompt>~.~\{text1\}~<mask>~,~\{text2\}} &  48/48/5793  \\
    SST-2   & 2  & single & \texttt{<prompt>~.~\{text\}~.~It was <mask>~.}   &  32/32/512   \\
    \bottomrule
\end{tabular}
\label{tab:task}
\end{table}

Our work is organized as follows. In Section 2, we summarize the related works. Section 3 details our methods that achieve competitive results. In Section 4, we present our experimental results. Finally we give an extended discussion of our work in Section 5.

\section{Related Work}

\subsection{Prompt-based methods}
Prompt-based methods perform zero-shot or few-shot learning in a natural way by using a template to transform the NLP task into a cloze task, and a verbalizer maps the predicted words into labels. These methods usually require carefully designed template and verbalizer to achieve comparable results to supervised methods. The template can be discrete, such as manually designed \citep{Brown2020LanguageMA, Schick2020AutomaticallyIW, Schick2020ExploitingCF}, mined from corpora \citep{Jiang2019HowCW}, or be continuous such as generative PTMs \citep{Gao2021MakingPL}, finetuned by back propagation \citep{Liu2021GPTUT} or be constructed using gradient-guided search \citep{Shin2020ElicitingKF}. There are also some works on improving verbalizer. \citeauthor{Cui2022PrototypicalVF} proposed the prototypical verbalizer which is built directly from training data and \citeauthor{Hu2021KnowledgeablePI} focused on incorporating external knowledge into the verbalizer, forming a knowledgeable prompt tuning (KPT), to improve and stabilize prompt tuning.

\subsection{Black-Box Tuning}
The Black-Box Tuning is proposed to optimize prompt under the Language-Model-as-a-Service scenario where the gradient cannot be accessed during the training process. Black-box Discrete Prompt Learning (BDPL) employed a variance reduced policy gradient estimator to approximate the gradients, and then update the prompts \citep{Diao2022BlackboxPL}.
BBT \citep{Sun2022BlackBoxTF} managed to optimize the continuous prompt prepended to the input text by iteratively querying the pretrained-model inference API. BBTv2 \citep{sun2022bbtv2} improved BBT with deep prompts that are injected to every layer of the pretrained-model.

\section{Methodology}

\subsection{Architecture}

\begin{figure}[!htb]
\centering
\includegraphics[width=\textwidth]{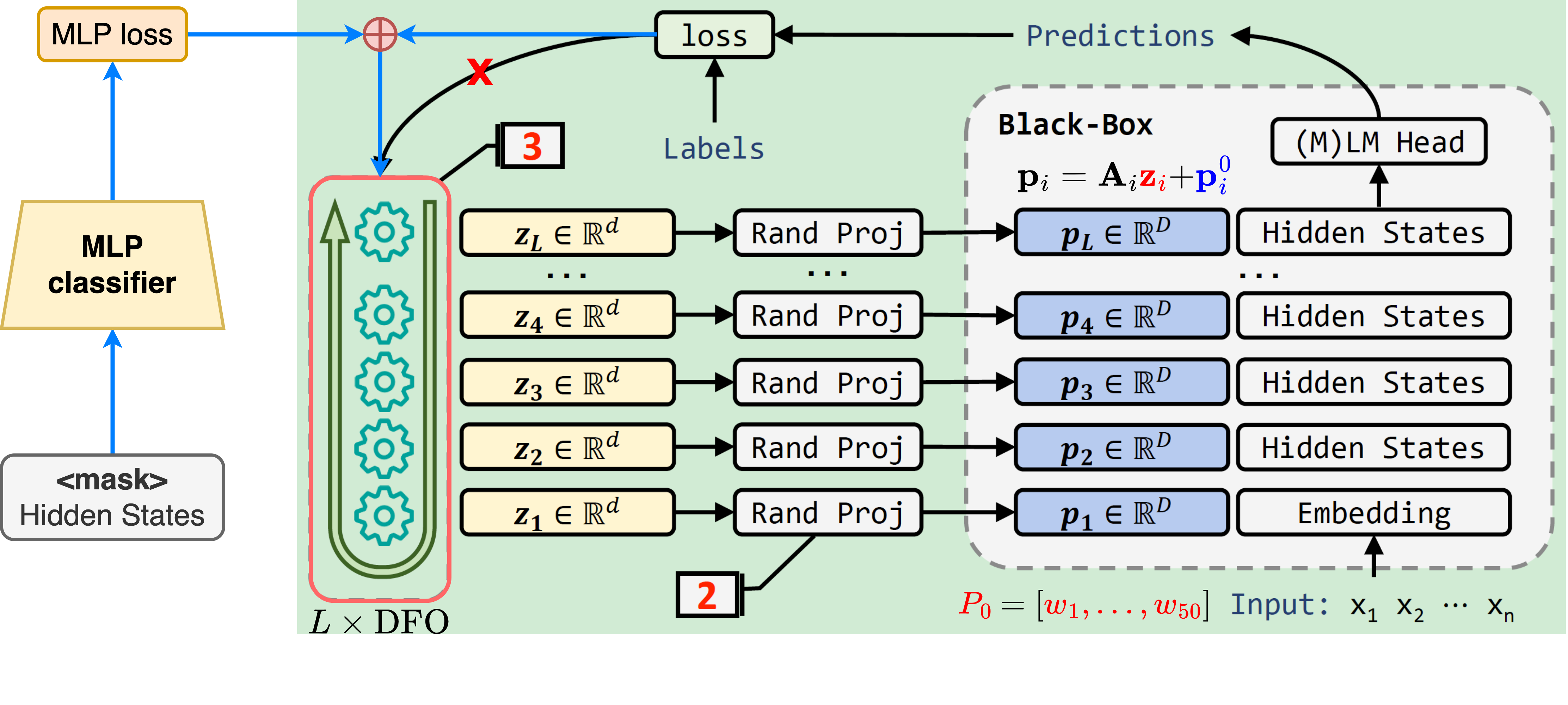}
\caption{Illustration of our solution based on BBTv2}
\label{fig:arch}
\end{figure}

The main architecture of our solution is presented as Figure \ref{fig:arch}.
We introduce several modifications to BBTv2 \citep{sun2022bbtv2}, which is the state-of-the-art black-box tuning method suitable for this competition.
BBTv2 adopts deep prompts by prepending different prompts to the hidden states at every layer of a given black-box pretrained model.
The prompts used by BBTv2 can be generated according to the following formulation.

\begin{equation}
\mathbf{p}_i = \mathbf{A}_i \mathbf{z}_i + \mathbf{p}_i^0, \qquad i = 1, 2, \dots, L
\label{eq:prompt}
\end{equation}
where $\mathbf{A}_i$ is a projection matrix at the $i$-th layer, staying fixed during training once initialized,
$\mathbf{z}_i$ is an intrinsic vector in a low-dimensional subspace to learn using some derivative-free optimization (DFO) algorithms,
$\mathbf{p}_i^0$ stands for an initial prompt embedding corresponding to the prompt tokens at $i$-th layer,
after feeding into the black-box model the initial prompt tokens $P_0 = [w_1, \dots, w_{50}]$ concatenated with training texts.
Basically, the learning of intrinsic vector $\mathbf{z}_i$ can be regarded as slightly tuning based on the initial prompt embedding $p_i^0$,
analog to the residual connection derived from ResNet \citep{he2016deep}.

In general, we come up with 8 modifications to BBTv2 method, 5 of which are proven to be effective meeting the requirements of this competition.
The 5 effective modifications are: multiple label words, selecting tokens as $P_0$, rolling strategy to optimize intrinsic vectors, multi-task loss from MLP classifier and classic ensemble.
The other three including curriculum learning using fake-labeled data, independent MLP with data augmentation and unsupervised calibration were not adopted in our final solution due to different reasons.
We present the details of all these modifications in following subsections.

\subsection{Multiple Label Words}
\label{sec:mlabel}

Following the methods of BBTv2, for text classification tasks, we can calculate the loss with output logits $\hat{y}$ over a candidate set of label words and the golden label word $\widetilde{y}$ on masked position.
Unlike BBTv2, each label may be mapped to multiple words, such as, we can map sentiment label ``positive'' to ``great'' and ``good''.
So we employ one-to-many label mapping in calculation that we select the word with the largest logit from candidate words of each label.
Specifically, Table \ref{multi-mapping} shows the label-words mapping for each task.

\begin{table}[!ht]
\caption{Label words of 5 tasks}
\centering
\begin{tabularx}{\textwidth}{c | X}
    \toprule
    Task    & Label Mappings  \\
    \midrule
    DBPedia & 0: Company/Corporation, 1: EducationalInstitution, 2: Artist, 3: Athlete/Sportsman, 4: OfficeHolder/Official, 5: MeanOfTransportation/Vehice/TrafficTransportation, 6: Building, 7: NaturalPlace, 8: Village, 9: Animal, 10: Plant, 11: Album, 12: Film, 13: WrittenWork \\
    QNLI    & 0: Yes, 1: No \\
    QQP     & 0: Yes, 1: No \\
    SNLI    & 0: Yes, 1: Maybe/Perhaps, 2: No \\
    SST-2   & 0: bad/awful, 1: great/good \\
    \bottomrule
\end{tabularx}
\label{multi-mapping}
\end{table}

\subsection{Selecting Tokens as \texorpdfstring{$P_0$}{P0}}
\label{sec:p0}

According to Eq (\ref{eq:prompt}), $\mathbf{p}_i^0$ transformed from the initial tokens of $P_0$ plays an important role in tuning deep prompts on a specific task.
A good choice of $P_0$ may ease the difficulty of outputing high probabilities over label words from \texttt{<mask>} token and hence accelerate the optimization of intrinsic vector $\mathbf{z}_i$ at each layer.
Driven by such a conjecture, we use an intuitive method to select tokens using training examples as $P_0$, different from randomly sampling using in BBTv2.

\begin{figure}[!htb]
\centering
\includegraphics[width=0.8\textwidth]{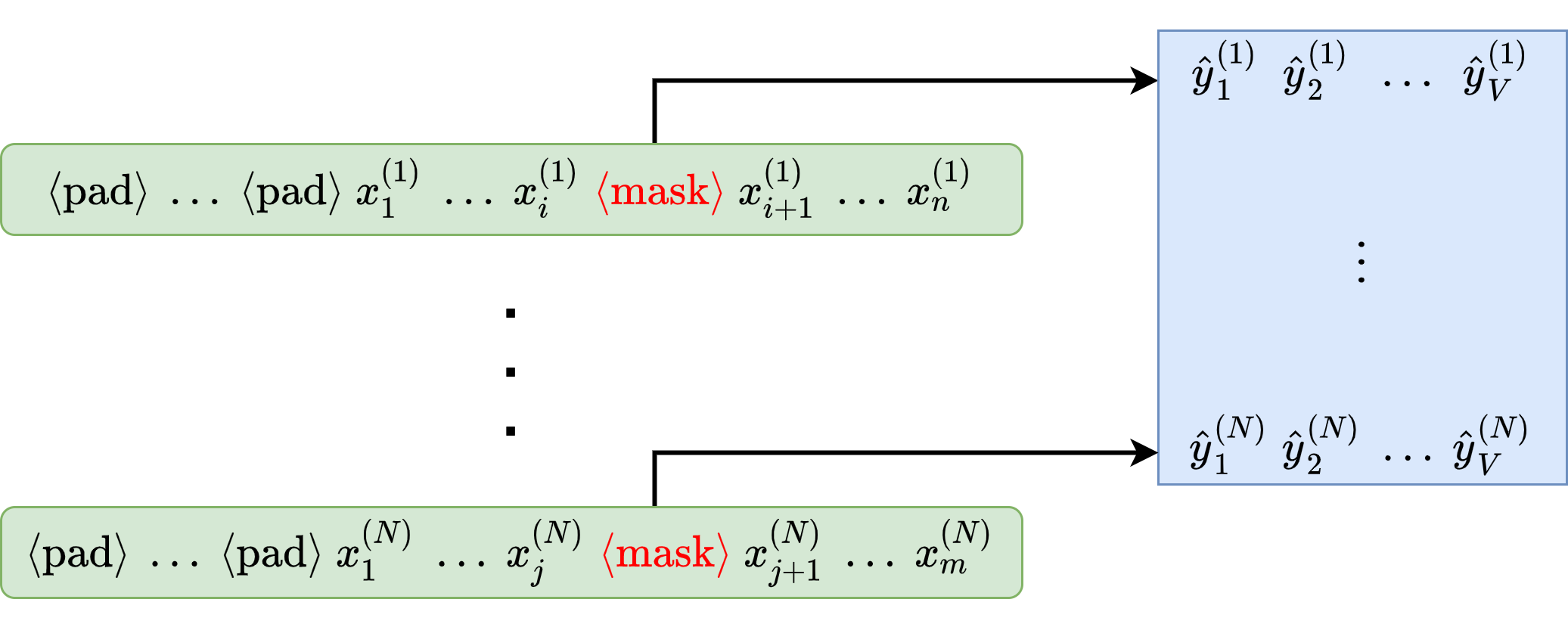}
\caption{Sample tokens from \texttt{<mask>} distributions}
\label{fig:samp}
\end{figure}

The selection process is conducted following the steps as below.
\begin{itemize}
    \item[1.] Use \texttt{<pad>} as initial prompts $P_0$, and feed into black-box model each training example by filling out the input template of a specific task.
    \item[2.] For each example, select 5000 ($100 \times |P_0|$) tokens with highest probabilities from the output distribution of \texttt{<mask>} position.
    \item[3.] For each selected token $w$ in the union set of step 2, compute the minimum probability $p_{min} (w)$ over all training examples.
        The probability is set to 1 by default if the token $w$ is not ranked in top 5000 for a given example.
    \item[4.] Select 50 tokens with most coverage over all examples and highest $p_{min}$.
        The coverage of a token is defined as the ratio of examples for which the token is ranked in top 5000 from corresponding distribution of \texttt{<mask>}.
\end{itemize}
We use the final 50 tokens as $P_0$ (Figure \ref{fig:samp}), which are much likely to be attended from the \texttt{<mask>} token in self-attention sublayers of the black-box model.

\begin{figure}[!htb]
\centering
\includegraphics[width=\textwidth]{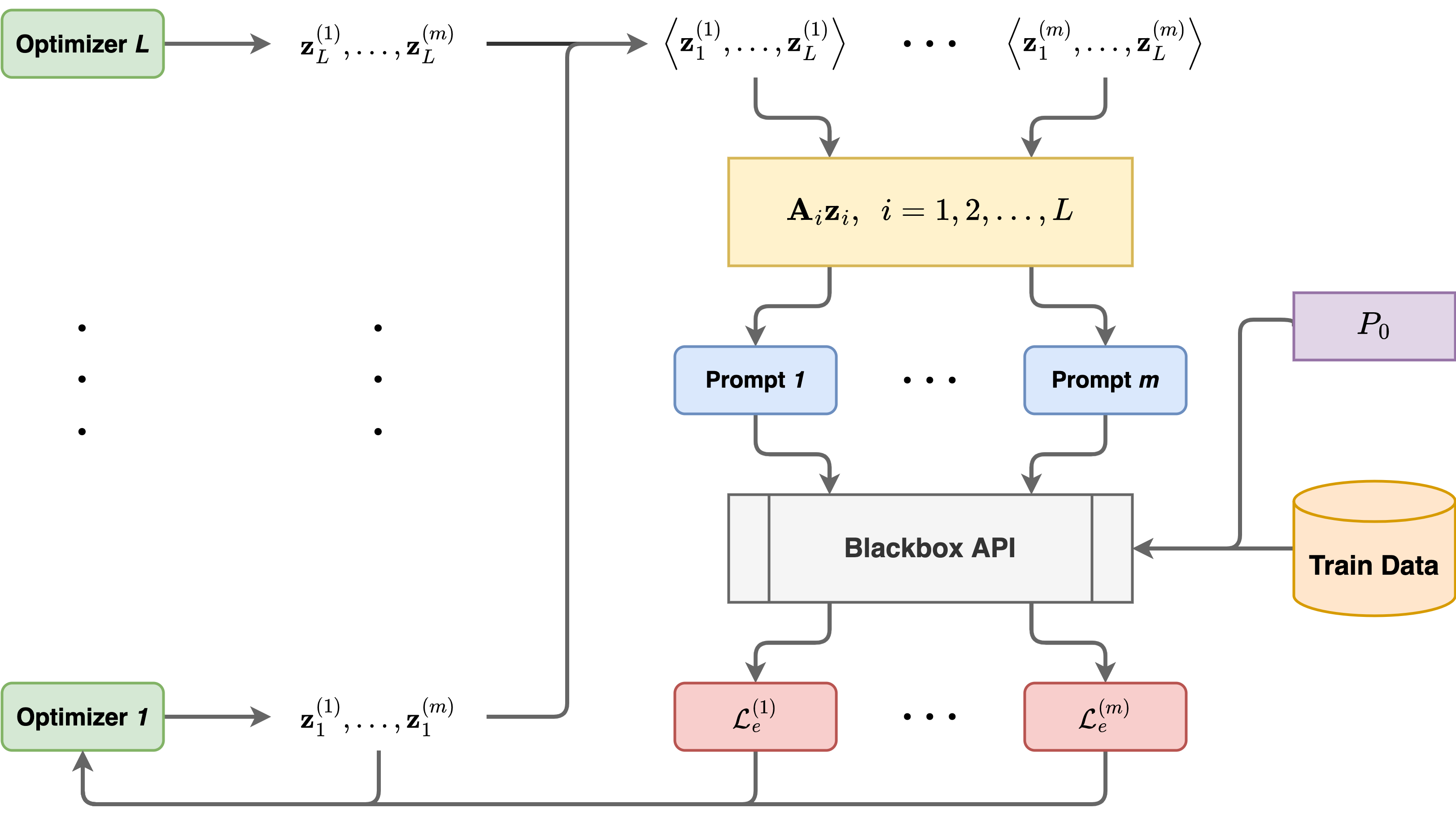}
\caption{Group and project intrinsic vectors into different sets of deep prompts, compute losses and update internal states of DFOs}
\label{fig:dfo}
\end{figure}

\subsection{Rolling Strategy to Optimize Intrinsic Vectors}
\label{sec:roll}

Despite of $P_0$, every intrinsic vector $\mathbf{z}_i$ requires an independent derivative-free optimizer, resulting in $L$ optimizers in total.
There are two major factors to dominate the training effect of all intrinsic vectors.
The first is the DFO algorithm to search best solutions of $\mathbf{z}$ which lead to appropriate deep prompts for a specific classification task.
After a thorough analysis, we choose CMAES (\textbf{C}ovariance \textbf{M}atrix \textbf{A}daptation \textbf{E}volution \textbf{S}trategy) same as BBTv2 \citep{hansen2016the,sun2022bbtv2}.
Another factor is the strategy used to optimize all intrinsic vectors. Since there are $L$ different intrinsic vectors and $L$ corresponding derivative-free optimizers,
the order of updating $L$ optimizers has a great impact on computational complexity and tuning performance.

We compare two ordinary strategies for such scenario and then propose a new rolling strategy instead.
The first is ``divide-and-conquer'' employed by BBTv2, according to which the intrinsic vector at each layer is optimized alternately using CMAES algorithm.
Namely, only one optimizer stays active to train corresponding intrinsic vector while the others are fixed at each iteration \citep{sun2022bbtv2} \footnote{\url{https://github.com/txsun1997/Black-Box-Tuning}}.
One complete iteration of CMAES optimizer consists of sampling $m$ (population size) solutions for an intrinsic vector and updating internal states using fitting errors corresponding to previous solutions.
An evitable consequence of this strategy is the significantly few updates of each optimizer, leading to insufficient learning of each intrinsic vector.
Given the total budget $B$ calling the black-box model, the number of updates is roughly $\frac{B}{L}$ for each optimizer.
Such consequence may become even worse when the black-box model scales up to larger size with much more Transformer layers.

\begin{algorithm}[!htb]
\caption{Sample solutions}
\begin{algorithmic}
\Require Iteration index $t \ge 0$, \\
    Derivative-free optimizers $\{\mathcal{M}_i\}_{i=1}^L$, \\
    Population size $m$, \\
    Number of \texttt{Unstable} optimizers $N_u$, \texttt{Testing} $N_t$, \texttt{Stable} $N_s$ ($N_u + N_t + N_s = L$)
\If{$t = 0$}  \Comment{Initialize states for all optimizers}
    \State Randomly set current state $S^i$ for each optimizer, making sure $S^1$ is \texttt{Unstable} for $\mathcal{M}_1$ \\
        \hskip 3em and the numbers of optimizers in different states equal to $N_u$, $N_t$ and $N_s$ respectively.
    \State Set last state $S_{old}^i$ of all optimizers to be \texttt{None}
    \State Set running threshold $\phi_i$ of all optimizers to be \texttt{INF}
\EndIf

\State $\mathcal{Z} \gets \{\}$
\For{$i = 1$ to $L$}
    \State $\mathbf{z}_{i, t}^{(k)} \gets \mathbf{0}, k = 1, \dots, m$  \Comment{Initialize solutions of current iteration}
    \If{$S^i$ is \texttt{Unstable}}
        \If{$S_{old}^i$ is \texttt{Unstable}}
            \State $\mathbf{z}_{i, t}^{(k)} \gets \mathbf{z}_{i, t - 1}^{(\delta_k)}$  \Comment{Shuffle $m$ existing solutions}
        \Else 
            \State $\mathbf{z}_{i, t}^{(k)} \gets \mathcal{M}_i$  \Comment{Sample new solutions from $\mathcal{M}_i$}
        \EndIf
    \EndIf
    \If{$S^i$ is \texttt{Testing} and $S_{old}^i$ is \texttt{Unstable}}
        \State $\mathcal{I} \gets \{k | \mathbf{z}_{i, t - 1}^{(k)} \text{ ranked in top } \frac{m}{2}\}$ \Comment{Select $\frac{m}{2}$ best solutions from last iteration}
        \State $\mathbf{z}_{i, t}^{(k)} \gets \mathbf{z}_{i, t - 1}^{(\delta_k)}, \ \delta_k \in \mathcal{I}$  \Comment{Randomly sample from $\frac{m}{2}$ solutions selected above}
        \State s.t. $\#\{k | \delta_k = j\} = 2, \ \forall j \in \mathcal{I}$  \Comment{Each of the selected solutions is exactly sampled twice}
    \EndIf
    \If{$S^i$ is \texttt{Stable} and $S_{old}^i$ is \texttt{Testing}}
        \State $\mathbf{z}_{i, t}^{(k)} \gets \mathbf{z}_{i, t - 1}^\ast$  \Comment{Reset with the best solution $\mathbf{z}_{i, t - 1}^\ast$ from last iteration}
    \EndIf
    \State $\mathcal{Z} \gets \mathcal{Z} \cup \left\{\mathbf{z}_{i, t}^{(1)}, \dots, \mathbf{z}_{i, t}^{(m)}\right\}$  \Comment{Save sampled solutions and random mappings $\delta_k$ if exists}
\EndFor

\State \Return $\mathcal{Z}$
\end{algorithmic}
\label{alg:stage1}
\end{algorithm}

The second strategy is to sample solutions and update internal states of all optimizers simultaneously in each iteration (referred to as ``all-in-time''), which was adopted in the first stage of this competition.
At first, $m$ solutions were generated as samples of $\mathbf{z}_i$ from the $i$-th optimizer.
Then $L \times m$ solutions were combined into $m$ groups randomly without replacement, and each group consists of only one solution from each optimizer (Figure \ref{fig:dfo}).
That is, the $i$-th solution $\mathbf{z}_i^{(j)}$ in the $j$-th group $[\mathbf{z}_1^{(j)}, \dots, \mathbf{z}_L^{(j)}]$ is randomly chosen from $m$ sampled solutions of the $i$-th optimizer.
Each group of solutions are projected into an ordered set of deep prompts by $\mathbf{A}_i \mathbf{z}_i$ at each layer.
By taking these deep prompts and all training examples as input to the black-box model, we can then compute a metric measuring the fitting error of the prompt.
We can choose any form of metrics, such as cross-entropy loss, hinge loss, inverse number of accuracy or F1 score and so on.
Every solution $\mathbf{z}_i^{(j)}$ in the $j$-th group is set to a same metric $\mathcal{L}_e^{(j)}$ as its error.
Gathering all solutions and their corresponding errors for a given optimizer, we can easily update its internal states.
Since all optimizers are updated in the end of an iteration, the computational complexity is approximately proportional to $\frac{BL}{m}$ if we omit some intricate changes of each CMAES optimizer due to different population size $m$.
Contrary to ``divide-and-conquer'' adopted in BBTv2, this ``all-in-time'' strategy enables each derivative-free optimizer to converge faster and better through sufficient iterations except the critical burden of computational hardware.

\begin{algorithm}[!htb]
\caption{Update internal states and ancillary attributes}
\begin{algorithmic}
\Require Iteration index $t \ge 0$, \\
    Derivative-free optimizers $\{\mathcal{M}_i\}_{i=1}^L$, \\
    Number of \texttt{Unstable} optimizers $N_u$, \texttt{Testing} $N_t$, \texttt{Stable} $N_s$ ($N_u + N_t + N_s = L$), \\
    Sampled solutions $\left<\mathbf{z}_{1, t}^{(1)}, \dots, \mathbf{z}_{L, t}^{(1)}\right>, \dots, \left<\mathbf{z}_{1, t}^{(m)}, \dots, \mathbf{z}_{L, t}^{(m)}\right>$, \\
    Fitting errors $\mathcal{L}_{e, t}^{(1)}, \dots, \mathcal{L}_{e, t}^{(m)}$
\For{$i = 1$ to $L$}
    \If{$S^i$ is \texttt{Unstable}}
        \State Compute average fitting errors for each solution with random mappings $\delta_k$
    \EndIf
    \If{$S^i$ is \texttt{Testing} and $t > 0$}
        \State Compute average fitting errors for each solution with random mappings $\delta_k$
        \State Update internal states of $\mathcal{M}_i$
    \EndIf
\EndFor

\If{$t = 0$}
    \State $\mathcal{I}_t \gets sample(\{i | S^i \text{ is \texttt{Unstable}}\},\ N_t)$
    \State $\mathcal{I}_s \gets sample(\{i | S^i \text{ is \texttt{Testing}}\},\ N_s)$
\Else
    \State $\mathcal{I}_t \gets sortby(\{i | S^i \text{ is \texttt{Unstable}}\},\ \#\{\mathcal{L}_e | \mathcal{L}_e \ge \phi_i\},\ N_t)$
    \State $\mathcal{I}_s \gets sortby(\{i | S^i \text{ is \texttt{Testing}}\},\ \| \mu^i - \mu_{old}^i \|,\ N_s)$
\EndIf

\For{$i = 1$ to $L$}  \Comment{Update ancillary attributes}
    \State $S_{old}^i \gets S^i$
    \State $S^i \gets \texttt{Unstable}$
    \If{$S_{old}^i$ is \texttt{Unstable}}
        \State $\phi_i \gets max(\phi_i, mean(\mathcal{L}_e))$
        \If{$i \in \mathcal{I}_t$}
            \State $S^i \gets \texttt{Testing}$
            \State $\phi_i \gets min(\phi_i, mean(\mathcal{L}_e))$
        \EndIf
    \EndIf
    \If{$S_{old}^i$ is \texttt{Testing}}
        \State $\phi_i \gets min(\phi_i, mean(\mathcal{L}_e))$
        \If{$i \in \mathcal{I}_s$}
            \State $S^i \gets \texttt{Stable}$
        \EndIf
    \EndIf
\EndFor

\State $t \gets t + 1$
\end{algorithmic}
\label{alg:stage2}
\end{algorithm}

\begin{figure}[!htb]
\centering
\includegraphics[width=0.85\textwidth]{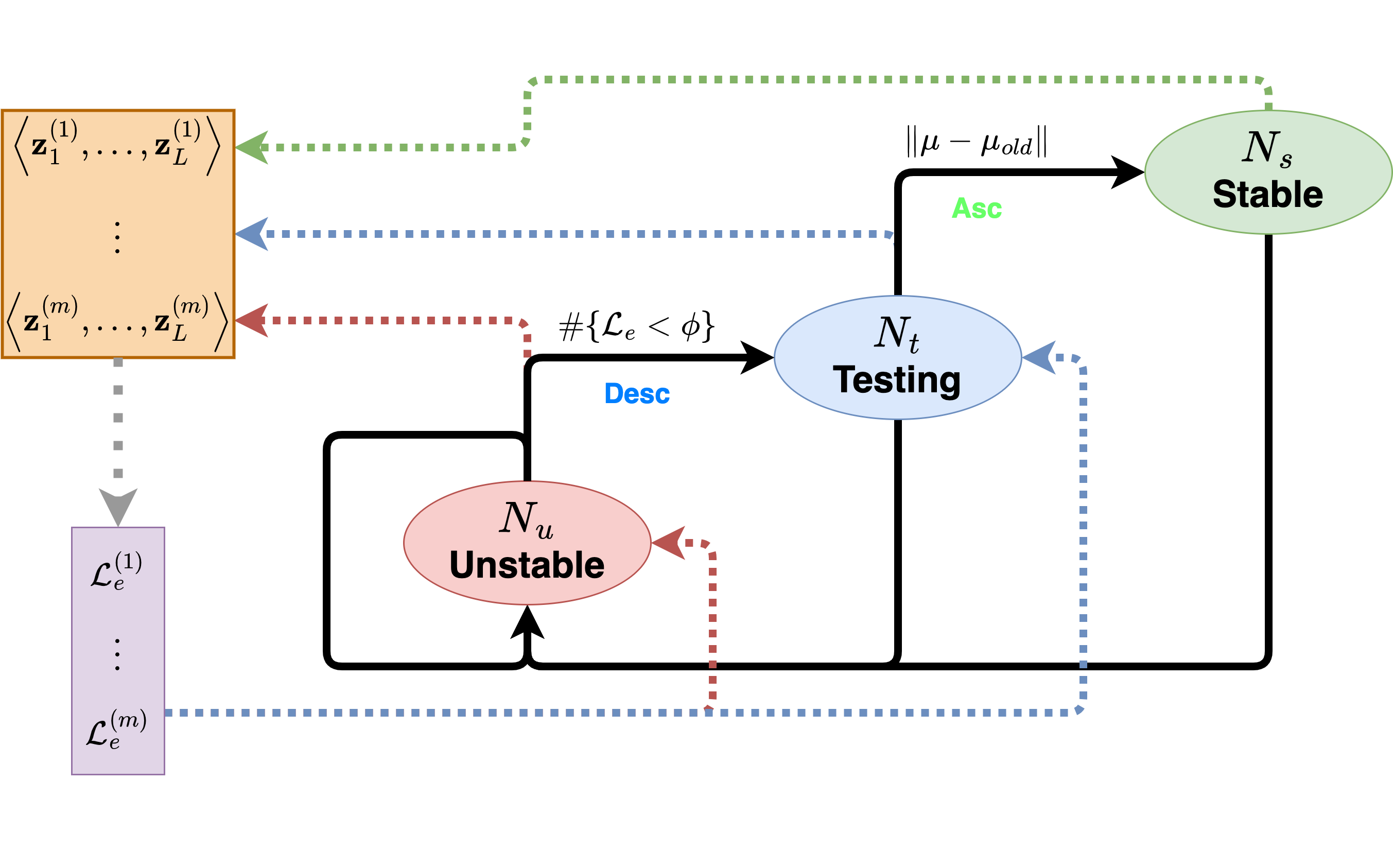}
\caption{State transition of derivative-free optimizers using ``rolling'' strategy}
\label{fig:roll}
\end{figure}

We then propose a novel ``rolling'' strategy inspired by the development standards of Debian softwares (Figure \ref{fig:roll}).
We define three states for each derivative-free optimizer: \texttt{Unstable}, \texttt{Testing} and \texttt{Stable}, as well as a running threshold $\phi$ for average fitting error.
The whole procedure of this ``rolling'' strategy consists of two alternating stages in each iteration, depicted in Algorithm \ref{alg:stage1} and \ref{alg:stage2}.
The first stage is to sample solutions as intrinsic vectors at each layer.
The sampled solutions are then combined and transformed into deep prompts as the process in Figure \ref{fig:dfo}, and fitting errors corresponding to this set of prompts are collected.
Using combined solutions and their corresponding errors, we perform the second stage to update internal states of each optimizer as well as some ancillary attributes.
The primary principle underlying this strategy is that, optimizers with better solutions are transferred from \texttt{Unstable} to \texttt{Testing} and optimizers with smaller updates are transferred from \texttt{Testing} to \texttt{Stable}.
By repeating this procedure, every optimizer gets updates in a ``rolling'' manner since it periodically enters \texttt{Testing} state.

In practice, we restrict $\frac{L}{2}$ optimizers in \texttt{Unstable} state, $\frac{L}{3}$ \texttt{Testing} and $\frac{L}{6}$ \texttt{Stable} at every iteration.
Note that the numbers of CMAES optimizers in different states can be adjusted according to computational performance of the local machine conducting the optimization.
Only one constraint needs to be satisfied, the numbers of \texttt{Unstable}, \texttt{Testing} and \texttt{Stable} optimizers should decrease monotonically, $N_u \ge N_t \ge N_s$.
We briefly illustrate the transition of different states in Figure \ref{fig:roll}.

\begin{table}[!htbp]
\centering
\caption{Complexity and number of updates per DFO using different strategies}
\begin{tabular}{c c c}
    \toprule
    & Complexity & Updates Per DFO \\
    \midrule
    divide-and-conquer & $\times \frac{B}{m}\:\:\:\:$ & $\frac{B}{m L}$      \\
    all-in-time        & $\times \frac{B L}{m}\:\:$   & $\frac{B}{m}$        \\
    rolling            & $\times \frac{B N_t}{m}$     & $\frac{B N_t}{m L}$  \\
    \bottomrule
\end{tabular}
\label{tab:dfo}
\end{table}

So far, three strategies have been discussed to facilitate the optimization of deep prompts for the black-box model.
The computational complexity and number of updates are roughly concluded in Table \ref{tab:dfo}.
Obviously, the ``rolling'' strategy trades off the computational complexity against sufficient updates given a small budget.

\subsection{Multi-task Loss from MLP Classifier}
\label{sec:mloss}

\begin{figure}[!htb]
\centering
\includegraphics[width=0.7\textwidth]{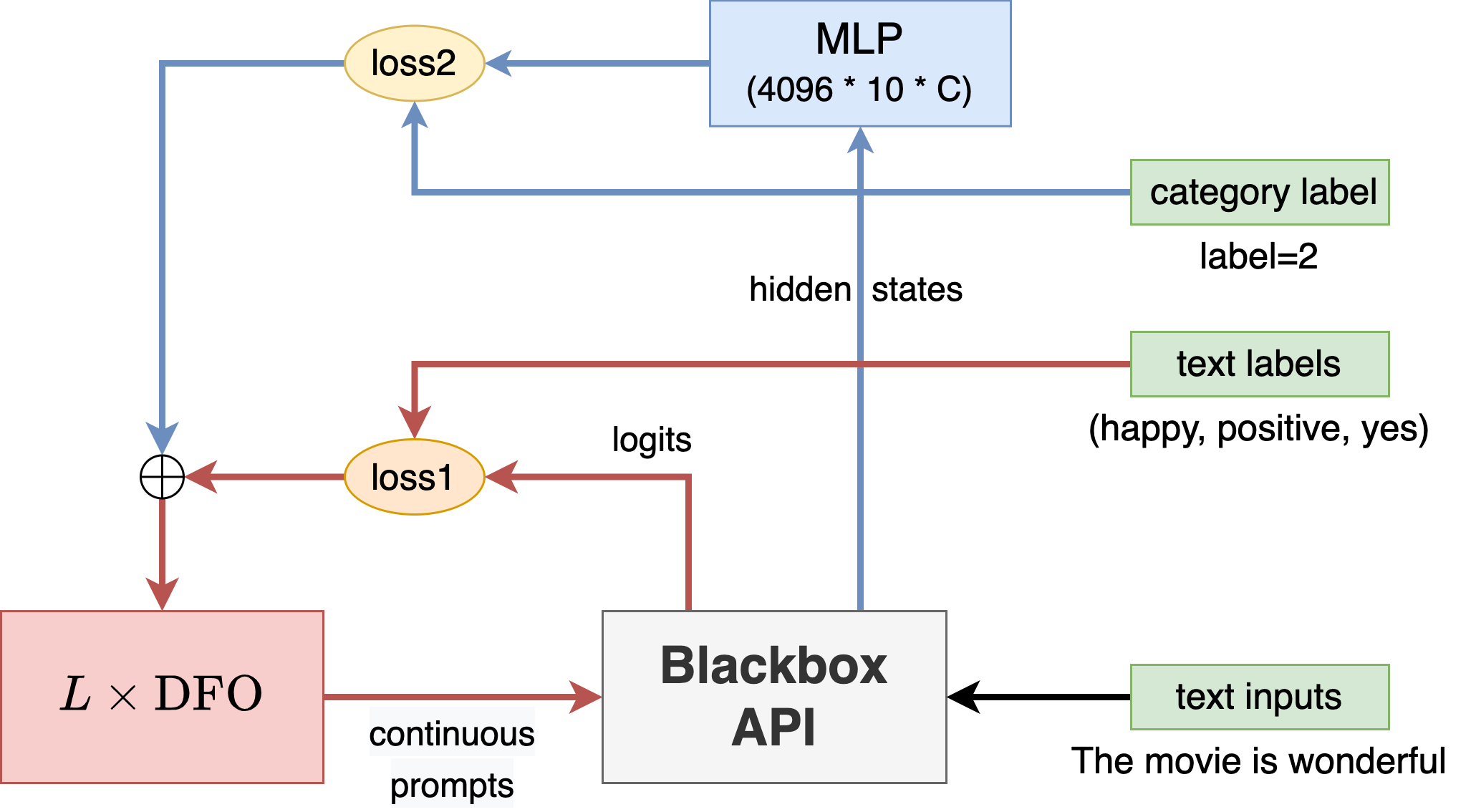}
\caption{Illustration of multi-task loss from MLP classifier}
\label{fig:mtask}
\end{figure}

Self-supervised language model is not designed for sementic classfication primordially.
The BBTv2 design a masked template, then take the generated text on the masked position as classification label.
According to our experiments, the precision is seriously enfluenced by the selection of template and label text.
In order to make it more robust, we directly categorize the hidden states of the \texttt{<mask>} token, instead of decoding it using learned token embeddings.
As shown in Figure 5, we concate last 4 layers hidden states of the \texttt{<mask>} token, then use a an multilayer perception (MLP) as classifier.
Finally, an 1-layer fully connected classifier is used, the activation function is sigmoid, the optimizer is Adam.
Since the number of training examples is quite small ($\le 112$) in the few-shot setting, we take the MLP classifier as a secondary training task to reduce the risk of overfitting.

To save computational cost, we introduce two training stages to the multi-task learning, iteratively switching between prompt updating and MLP updating.
Firstly we fix the MLP parameters, and calculate the total loss to update prompt with DFO algorithm.
Total loss is composed of two parts, loss1 and loss2 (Figure \ref{fig:mtask}).
Where loss1 is cross-entropy loss of language model's logits and text labesl, and loss2 is cross-entropy loss of MLP's logits and category label.
Secondly we fix the prompt, and update the MLP parameters with gradiant algorithm.
Alternately repeat these two training stage.
Concretely, we update the MLP once after updating prompt every 100 steps.

\subsection{Ensemble}
\label{sec:ensem}

Besides four modifications presented above, we also employ a famous technique, ensemble learning, to improve the generalization capability of learned prompts.
After a thorough analysis of different ensemble methods including Bagging, SAMME, and SAMME.R, we finally choose naive ``Top K'' to fuse the predictions of K best prompts or MLP classifiers.
The way to generate the final predictions is formulated as follows
\begin{equation}
    \hat{y} = \arg \max_{c \in \{1, \dots, C\}} \frac{1}{\| \mathcal{P} \|} \sum_{\mathbf{p} \in \mathcal{P}} f_{BBT}(y = c | \mathbf{x}, \mathbf{p})
\label{eq:ensem}
\end{equation}
where $\mathcal{P}$ is a set of saved deep prompts or MLP classifiers with corresponding prompts, $C$ is the number of classes of a specific task, $f_{BBT}(\cdot)$ stands for probability distribution over $C$ classes using the black-box model with a set of deep prompts or a MLP classifier.
In practice, we save 3 best sets of deep prompts and 3 best MLP classifiers with prepended prompts, resulting in 6 predictions to fuse for every example.

\begin{figure}[!htb]
\centering
\includegraphics[width=0.6\textwidth]{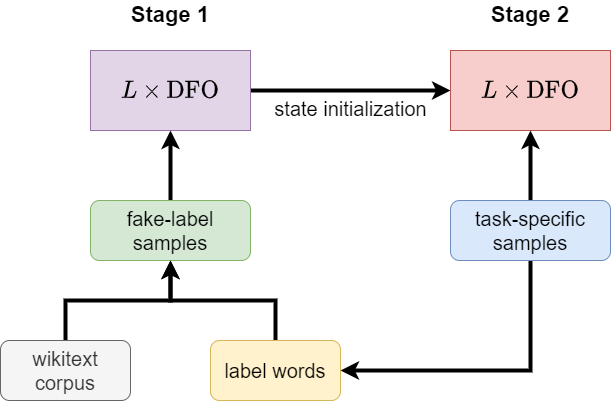}
\caption{Two-stage curriculum learning using fake-labeled data}
\label{fig:fake}
\end{figure}

\subsection{Curriculum Learning using Fake-labeled Data}
\label{sec:fake}

In order to train the optimizer more stable and improve the generalization of the prompt, we explore the curriculum learning with unlabeled data,
since there are only dozens of training examples for each task.

\begin{table}[!htb]
\centering
\caption{Result of curriculum learning with fake-labeled data}
\begin{tabular}{c c c}
    \toprule
    \multirow{2}{*}{Task} & Baseline & Curriculum Learning \\
    & (Test F1) & (Test F1)  \\
    \midrule
    DBPedia & 0.9297 & 0.9221 \\
    QNLI    & 0.7015 & 0.6802 \\
    QQP     & 0.6093 & 0.6542 \\
    SNLI    & 0.5032 & 0.5174 \\
    SST-2   & 0.8716 & 0.8854 \\
    \midrule
    Avg     & 0.7231 & 0.7319 \\
    \bottomrule
\end{tabular}
\label{table:fake}
\end{table}

Based on wikitext corpus \cite{merity2016pointer} and label words of all tasks, we construct some pseudo-label samples.
Specifically, if a label word appears in the text of wikitext, we replace it with the \texttt{<mask>} token, and predict the real label word during training process.
The process of curriculum learning is shown in the Figure \ref{fig:fake}.
In the first stage, we first train all derivative-free optimizers $B_{fake}$ steps with $N_{fake}$ pseudo-label examples; then we continue to train the optimizers using task-specific training examples.
We set $N_{fake} = 6000$ and $B_{fake}=1000$, the experimental results are presented in Table \ref{table:fake}.
There are significant improvements for QQP, SNLI and SST-2, and each task takes 5 more hours for training.

\subsection{Independent MLP with Data Augmentation}
\label{sec:damlp}

Suffering from very few training examples, we investigate diverse data augmentation methods to enlarge the training datasets.
Following the method proposed in \citet{wei-zou-2019-eda}, we expect to employ four operations to augment training examples: synonym replacement (SR), random insertion (RI), random swap (RS) and random deletion (RD) \footnote{\url{https://github.com/jasonwei20/eda_nlp}}.
However, only RI and RS exhibit significant improvements beyond random fluctuation of testing performance on five classification tasks.
We eventually apply RI and RS to expand training examples by 6 times, and development examples by 2 times.
Due to enormous increasing of training time when augmenting examples in the optimization of deep prompts, we propose to train an independent MLP classifier using hidden states of the black-box model with augmented data.
Given a set of deep prompts with $L$ components at each layer of the black-box model, we concatenate the hidden states of last 4 layers from \texttt{<mask>} position as input embedding of an example.
It is apparent that we can collect 10 embeddings from a given example using 10 different sets of deep prompts saved in the derivative-free optimization.
Finally, input instances to train a 3-layer fully connected classifier are enlarged by approximately 60 times through data augmentation and using 10 best sets of deep prompts.
We use an AdamW optimizer to train this classifier with 100 epochs and save the best checkpoint on the development instances.
It is noteworthy that training MLP classifier with expanded instances derived from other 9 sets of deep prompts also tends to force the classifier learning from actual texts rather than optimized prompts.
The predictions of this MLP classifier can be fused as Eq (\ref{eq:ensem}).

\subsection{Unsupervised Clustering}
\label{sec:unsup}
We draw inspiration from recent findings \citep{Aharoni2020UnsupervisedDC} that texts in the same domain tend to be clustered together in the PLM embedding space, and thus can be used for classification. We attempted to train a GMM model based on the last hidden states of \texttt{<mask>} token. The experimental results revealed that this strategy improved greatly on DBPedia, QQP and SST-2 datasets, but declined significantly on the SNLI and QNLI datasets, and the overall result was the same as the baseline.This method was not used in the final submission, but it is still worth further research to find out why it performs so differently on different datasets, which will be covered in the discussion.

\begin{table}[!htb]
\centering
\caption{Ablation results of every modifications}
\begin{tabular}{l | c | c | l}
    \toprule
    & Performance (\%) $\uparrow$  & Time Cost $\downarrow$  & Comments  \\
    \midrule
    Multiple Label Words (\ref{sec:mlabel}) & single +2.73 & +0 & \\ \hline
    Selecting Tokens as $P_0$ (\ref{sec:p0}) & single +4.22 & +0 & \\ \hline
    Rolling Strategy (\ref{sec:roll}) & single +0.30 & -8hr & Compared to ``all-in-time'' \\ \hline
    \multirow{2}{*}{Multi-task Loss (\ref{sec:mloss})} & DFO single +1.10 & \multirow{2}{*}{+0} & \\
    & MLP single +0.25 & & \\ \hline
    Ensemble (\ref{sec:ensem}) & ensemble +2.46 & +0 & \\ \hline\hline
    Curriculum Learning (\ref{sec:fake}) & single +1.00 & +5hr & Great time cost \\ \hline
    Independent MLP (\ref{sec:damlp}) & ensemble +0.51 & +0.7hr & No optimization of prompts \\ \hline
    \multirow{2}{*}{Unsupervised Clustering (\ref{sec:unsup})} & single +0.01 & \multirow{2}{*}{+0} & \multirow{2}{*}{Bad at QNLI \& SNLI} \\
    & QNLI -2.91 SNLI -1.78 & & \\
    \bottomrule
\end{tabular}
\label{tab:abl}
\end{table}

\section{Experiments}

We conclude the results of previous method in this section.
Since we experimented so many modifications during a short competition cycle, there is no guarantee that all experiments share same hyper-parameters, controlled configurations and hardware conditions.
Therefore, a rough ablation study is conducted under non-strictly consistent conditions.
Namely, we only guarantee the conditions experimenting with or without a given modification were kept the same, but differed across experiments on different modifications.
The analysis is summarized in Table \ref{tab:abl}.

Our primary modifications explicitly improve the tuning performance of deep prompts, including multiple label words, selection of $P_0$ tokens, rolling strategy, multi-task loss from MLP and ensemble.
Compared to ``all-in-time'' strategy, our proposed ``rolling'' strategy achieve slightly better performance (+0.3\%) and significantly decrease the learning cost of multiple derivative-free optimizers (-8 hours).
Except five effective modifications, we discard three other improvements in the final solution for different reasons.
Optimizing deep prompts using two-phase curriculum learning with fake-labeled data greatly increases training time, making this method uneconomic for just 1\% improvement.
Training independent MLP classifier with augmented datasets, on the other hand, has nothing to do with prompt tuning although it can boost final performance with little time cost.
This gradient-based method essentially violates the primary request of this competition and thus is abandoned.
The idea of calibrating predictions using unsupervised clustering fails in this situation due to significant performance degradation in QNLI and SNLI tasks.
At last, only top 5 modifications in Table \ref{tab:abl} were included in our submitted solution and produce results on five classification tasks.

\section{Discussion}

\subsection{Issues of SNLI Datasets}

\begin{table}[!htb]
\centering
\caption{Critical words affected by NLTK tokenization}
\begin{tabular}{l | l | l | l}
    \toprule
    \multirow{2}{*}{Original Words} & Words w/ & \multirow{2}{*}{Original Encodings} & Encodings w/  \\
    & NLTK Tokenization & & NLTK Tokenization  \\
    \midrule
    \texttt{won\textrm{\textquotesingle}t}  & \texttt{wo n\textrm{\textquotesingle}t}    & [351, 75]       & [19958, 295, 75]      \\
    \texttt{isn\textrm{\textquotesingle}t}  & \texttt{is n\textrm{\textquotesingle}t}    & [965, 75]       & [16, 295, 75]         \\
    \texttt{didn\textrm{\textquotesingle}t} & \texttt{did n\textrm{\textquotesingle}t}   & [399, 75]       & [222, 295, 75]        \\
    \texttt{\textrm{"}AND\textrm{"}}        & \texttt{`` AND ''}                         & [22, 5945, 113] & [45518, 4248, 12801]  \\
    \texttt{he\textrm{\textquotesingle}s}   & \texttt{he \textrm{\textquotesingle}s}     & [37, 18]        & [37, 128, 29]         \\
    \texttt{chef\textrm{\textquotesingle}s} & \texttt{chef \textrm{\textquotesingle}s}   & [8172, 18]      & [8172, 128, 29]       \\
    \bottomrule
\end{tabular}
\label{tab:dword}
\end{table}

After a careful check, we found crucial difference between training/development and test datasets of all seeds.
Clearly, word tokenization were performed only on training/development datasets but not on test \footnote{\url{https://github.com/jubgjf/PLMTuningCompetition}}.
By comparing the original examples of SNLI to the competition datasets, we uncover the tokenization operation described as \texttt{\color{red}{" ".join(nltk.word\_tokenize(text))}}.
Although most of words remain the same after NLTK tokenization, there are still significant changes in output encodings from RoBERTa tokenizer.
Table \ref{tab:dword} shows several words of great importance to entailment classification of text pairs, which differ a lot after word tokenization.
Negative words like \texttt{won\textrm{\textquotesingle}t} and \texttt{didn\textrm{\textquotesingle}t} basically determine the entailment relations of text pairs in most cases.
Such discrepancy induced from NLTK tokenization certainly destroys the generalization ability of deep prompts learned from biased training datasets since critial patterns like \texttt{n\textrm{\textquotesingle}t} are missing in SNLI test dataset.

\begin{table}[!htb]
\centering
\caption{Influence of NLTK tokenization on input encodings and predictions}
\begin{tabular}{c | c | c | c}
    \toprule
    \multirow{2}{*}{seed} & Encoding Changes of & Encoding Changes of & Prediction Changes of  \\
    & Training Examples (\%) & Development Examples (\%) & Test Dataset (\%)  \\
    \midrule
    13  & 70.83  & 79.17  & 53.98  \\
    42  & 77.08  & 79.17  & 45.24  \\
    50  & 60.42  & 85.42  & 16.38  \\
    60  & 79.17  & 79.17  & 24.65  \\
    8   & 85.42  & 83.33  & 27.00  \\
    \midrule
    Avg & 74.58  & 81.25  & 33.45  \\
    \bottomrule
\end{tabular}
\label{tab:dstat}
\end{table}

Besides some evident cases, we also evaluate the impact of NLTK tokenization on the input and output of our trained prompts (Table \ref{tab:dstat}).
In the few-shot setting of this competition, minor changes in a few words actually lead to huge differences of input encodings.
As in Table \ref{tab:dstat}, nearly 75\% of all encoding sequences are affected in training/development dataset of each seed.
From the viewpoint of prediction, 1/3 of final results become different if we apply the same deep prompts with test dataset preprocessed by NLTK tokenization.
Therefore, we argue that such preprocessing on only training/development datasets is unjustified from a practical perspective.

\begin{table}[!htb]
\centering
\caption{Performance of different solutions on SNLI datasets}
\begin{threeparttable}
\begin{tabular}{c || c | c || c | c || c | c | c || c | c}
    \toprule
    \multirow{2}{*}{seed} & \multicolumn{2}{c||}{First Stage} & \multicolumn{2}{c||}{Train/Dev/Test} & 
    \multicolumn{3}{c||}{Train$^\dag$/Dev$^\dag$/Test} & \multicolumn{2}{c}{Train$^\dag$/Dev$^\dag$/Test$^\dag$} \\ \cline{2-10}
    & Ours & BBTv2 & Ours & BBTv2 & Ours & BBTv2 & BBTv1 & Ours & BBTv2 \\
    \midrule
    13  & 0.6374  & 0.5661  & 0.4489  & 0.4327  & 0.4464  & 0.2974  & 0.3734  & 0.5422  & 0.4146  \\
    42  & 0.6814  & 0.5653  & 0.5160  & 0.4558  & 0.2543  & 0.3793  & 0.4096  & 0.5621  & 0.4071  \\
    50  & 0.6937  & 0.4872  & 0.4919  & 0.4411  & 0.2541  & 0.3294  & 0.3456  & 0.5263  & 0.4182  \\
    60  & 0.6992  & 0.5051  & 0.4989  & 0.4357  & 0.4686  & 0.3886  & 0.3445  & 0.4990  & 0.4116  \\
    8   & 0.7189  & 0.568   & 0.4618  & 0.4321  & 0.4494  & 0.3932  & 0.3283  & 0.5740  & 0.3766  \\
    \midrule
    Avg & 0.6861  & 0.5383  & 0.4835  & 0.4395  & 0.3746  & 0.3576  & 0.3603  & 0.5407  & 0.4056  \\
    \bottomrule
\end{tabular}
\begin{tablenotes}
    \small
    \item $^\dag$ Texts are preprocessed by NLTK tokenization.
\end{tablenotes}
\end{threeparttable}
\label{tab:dmodel}
\end{table}

In spite of the biased SNLI datasets, we further investigate the effectiveness of our proposed solution.
According to Table \ref{tab:dmodel}, our solution significantly surpasses the classic BBTv1 and BBTv2 even through using the biased datasets.
We consider two ways to eliminate the bias from training/development or test datasets.
If we optimize deep prompts using original training/development datasets without NLTK tokenization, the F1 score on test dataset goes from 0.3746 to 0.4835.
Other than that, we achieve much better score 0.5407 if applying same tokenization on test texts as training/development datasets.
To sum up, our solution surpasses classic black-box tuning methods like BBTv1 or BBTv2 to a great extent, and consequently we would achieve a much better rank in this competition if not affected by the biased SNLI task.

\subsection{Concerns about this Competition}

Although this competition is over, we still have several concerns about the procedure of organization and verification of proposed solutions of all teams for further discussion.
For simplicity, we just present our suggestions to some non-standard or controversial behaviors as follows.
\begin{itemize}
\item[1.] No public leaderboard in final stage.
    Since public evaluation of predicted results were cancelled on purpose in final stage, all teams have no access to ensure the correctness of proposed solutions and even the validity of encrypted test datasets.
Moreover, unexpected factors like minor code errors, random noises and runtime circumstance probably become dominate in the final predictions and ranks of some teams.
\item[2.] Lack of universality in terms of black-box models and tasks.
    Although the committee claims that final solutions of all teams should be universal and applicable to other classification tasks using distinct pretrained models, the black-box model and two of five tasks including random seeds and test datasets remain the same as the first stage.
It is highly possible that proposed methods of some teams might perform worse on other tasks using different pretrained models.
\item[3.] No statement or settlement about the questionable SNLI datasets.
    The evitable issue about biased SNLI datasets were not properly addressed even though we report our observation and analysis completely to the committee.
To our best knowledge, there is no analogous situation in real-world applications that only training/development datasets were preprocessed through some cleaning tricks rather than test texts, especially in a few-shot setting and prompt tuning.
\item[4.] Careless verification of proposed solutions.
    Although the committee is responsible enough to reproduce final predictions using submitted solutions, the validity of these solutions may not be carefully verfied.
The primary request of this competition is to optimize continuous or discrete \textbf{prompts} for each task.
However, solutions based on gradient optimization like independent MLP training and model distillation were clearly forbidden although they have no concern or little concern with prompt tuning.
Intuitively, such solutions violating primary request of prompt tuning have great potential to accomplish good results by extracting massive knowledge of the black-box model from hidden states of each layer.
\end{itemize}
We hope these flaws could be paid enough attention by the committee of this competition and conduct fair and just competitions in the future.

\section{Conclusion}

In this report, we summarize the details of our proposed solution for the competition ``Prompt Tuning using Pretrained Language Model''.
On the basis of BBTv2, we introduce five modifications bringing significant improvements to our final solution, including multiple label words, selection of $P_0$ tokens, ``rolling'' strategy, multi-task loss from MLP classifier and classific ensemble method.
We also present three more modifications with different shortcomings for this competition.
Finally, we discover the technical bias in SNLI datasets and analyze our solution in such unfair case.

\bibliography{anthology,custom}

\end{document}